# Text Classification based on Multiple Block Convolutional Highways

Seyed Mahdi Rezaeinia [a,b,*], Ali Ghodsi [a], Rouhollah Rahmani [b]

smrezaei@uwaterloo.ca

[a] Department of Statistics and Actuarial Science, University of Waterloo, Waterloo, Canada
[b] Network Science and Technology Department, University of Tehran, Tehran, Iran

## Abstract

In the Text Classification areas of Sentiment Analysis, Subjectivity/Objectivity Analysis, and Opinion Polarity, Convolutional Neural Networks (CNN's) have gained special attention because of their performance and accuracy. In this work, we applied recent advances in CNN's and propose a novel architecture, Multiple Block Convolutional Highways (MBCH), which achieves improved accuracy on multiple popular benchmark datasets, compared to previous architectures. The MBCH is based on new techniques and architectures including highway networks, DenseNet, batch normalization and bottleneck layers. In addition, to cope with the limitations of existing pre-trained word vectors which are used as inputs for the CNN, we propose a novel method, Improved Word Vectors (IWV). The IWV improves the accuracy of CNN's which are used for text classification tasks.

## 1. Introduction

Text classification or text categorization is an important Natural Language Processing (NLP) task and it takes an increasing interest in the research community. Text classification has various applications in the form of subject categorization, sentiment analysis, news aggregation, web search, spam detection and etc. Machine learning and word embeddings are crucial ingredients of text classification tasks.

There are different machine learning techniques for text categorization tasks such as decision tree, naive Bayes, k-nearest neighbor and support vector machines, but recently deep learning models such as Recurrent Neural Networks (RNN's), Long Short-Term Memory (LSTM) and Convolutional Neural Networks (CNN's) have drawn much attention in the field of NLP.

Among mentioned deep learning models, CNN's have been used in various NLP tasks such as opinion polarity detection and sentiment analysis [1,2,3], machine translation [4,5] language modeling [6,7], question answering [8,9,10] and other traditional NLP tasks [11]. Also, recent studies have shown that CNN's have notable performance compared to other Deep Neural Networks (DNN's), Gaussian Mixture Model and Hidden Markov Model (GMM-HMM) methods [12,13]. Gu et al. [14] pointed out that CNN's achieve better classification accuracy than other methods especially, on large datasets. Also, CNN's have impressive results in automatic speech recognition [15]. According to Bhushan et al. [16], CNN's perform better than recursive neural networks in capturing semantic information of the text. In addition, Gehring et al. [5] believe that convolutional encoders perform better or equal than the bi-directional LSTM encoders. Dauphin et al. [7] found that gated convolutional neural networks have 20x higher responsiveness than LSTMs.

To achieve higher accuracy, some authors have used very deep convolutional neural networks. For example, Conneau et al. [17] presented very deep CNN's for text classification. They showed the performance of their proposed model increases with the depth. Huang et al. [18] proposed a deep convolutional neural network architecture for object recognition task. In their architecture, each layer uses feature maps of all preceding layers as inputs. Zhang et al. [19] applied very deep CNN's for end-to-end speech recognition and achieved a large error rate reduction. However, very deep neural networks are time-consuming and suffer from vanishing and exploding gradient problems [20,21]. Highway network, which was introduced by Srivastava et al. [22] is an appropriate way to overcome these problems. By using highways, it is possible to train very deep neural networks and optimize the networks with virtually arbitrary depth. Another advantage of the highway networks is that increased flexibility in controlling how components of the input are carried. This flexibility improves the performance of deep and shallow neural networks [23].

Word embedding is another way to improve the performance of text classification based on deep neural networks. Word Embedding is one of the most useful deep learning methods used for constructing vector representations of words and documents. Word2Vec [24,25] and Global Vectors (GloVe) [26] are currently among the most accurate and usable word embedding methods which can convert words into meaningful vectors. These methods have achieved a lot of attention in text categorization because of their abilities to capture the syntactic and semantic relations among words. However, these methods have several limits and need to be improved. The Word2Vec and GloVe need very large corpora for training and generating exact vectors which are used as inputs of deep learning models [27,28]. For instance, Google has used about 100 billion words for training Word2Vec algorithms and has re-released pre-trained word vectors with 300 dimensions. Because of the small size of some datasets, researchers have to use pre-trained word vectors such as Word2Vec and GloVe, which may not be the best fit for their data. [1,29,30,31,32,33]. Another problem of Word2Vec and GloVe techniques is ignoring the sentiment information of the given text

[28,34,27]. Also, word vector calculations of the two methods that are used to represent a document do not consider the context of the document [35].

In this paper, we make use of recent advances in CNN's [22,36,37,18] and propose a novel architecture, Multiple Block Convolutional Highways (MBCH), for text classification tasks. We explored and applied some of these techniques such as highway network [22], DenseNet [18], batch normalization [37] and bottleneck layers [36,38]. The MBCH is a shallow network but we achieved state-of-art results on different benchmark datasets. In addition, in order to cope with the limitations of existing pre-trained word vectors, we propose a novel method, Improved Word Vectors (IWV), which improves the accuracy of pre-trained Word2vec word embeddings. The organization of this paper is as follows: Section 2 describes the related works and literature review for this research. Section 3 presents our proposed architecture and method. Section 4 reports our experiments, showing results along with evaluations and discussions. Section 5 is the conclusion of this research.

## 2. Related works

In this section, we review the related work from the following three perspectives: word embeddings, convolutional neural networks and highway networks.

### 2.1 Word embeddings

The use of word embeddings for text classification tasks has become a standard approach. Deep learning based text classification gives a vectorized value to a word using word embeddings or statistical methods. Each dimension of the generated vectors encodes a different aspect of words. Most of the deep learning tasks in natural language processing have been oriented towards methods which using word vector representations [28]. Word2Vec and GloVe are two successful word embedding algorithms which have high accuracy. Both methods are continuous vector representations of words algorithms and are very useful in text classification, clustering and information retrieval. In addition, they have some advantages compared to bag-of-words representation. For example, words close in meaning are near together in the word embedding space. In addition, Also, word embeddings have lower dimensionality than the bag-of-words [24].

However, the accuracy of the Word2vec and GloVe word embeddings depends on text corpus size. Meaning, the accuracy improves with the growth of text corpora. For instance, Severyn and Moschitti [39] applied Word2Vec method to learn the word embeddings on 50M tweets and used generated pre-trained vectors as inputs of a deep learning model. Qin et al. [40] have trained Word2Vec algorithm by English

Wikipedia corpus which has 408 million words. Because of the limitations and restrictions in some corpora, investigators prefer to use pre-trained word embeddings vectors as inputs of machine learning models. Kim [1] used pre-trained Word2Vec vectors as inputs to convolutional neural networks and increased the accuracy of text classification. Caliskan et al. [31] have used pre-trained GloVe vectors for increasing the accuracy of their proposed method. Also, Camacho-Collados et al. [30] utilized pre-trained Word2Vec vectors for the representation of concepts.

## 2.2 Convolutional neural networks

Convolutional Neural Networks have lately received great attention because of their state-of-the-art performance in all fields of NLP and computer vision. Kim [1] proposed a basic and effective multichannel CNN model for text categorization tasks which achieved high performance in sentiment analysis. According to his research, multiple convolutional layers can extract high-level abstract features. Kalchbrenner et al. [6] introduced a Dynamic Convolutional Neural Network (DCNN) architecture which uses dynamic k-max pooling for sentence modeling. Zeng et al. [41] used a deep convolutional neural network to extract lexical and sentence level features relation classification. Kotzias et al. [42] have proposed a general framework based on CNN's for polarity prediction of three review datasets. Similarly, Tang et al. [43] introduced a new method for sentiment classification and used product and user information as the input of a CNN. Zhang et al. [44] used character-level convolutional networks (ConvNets) for text classification. They used characters instead of words as input and the model contained 3 fully-connected layers and 6 convolutional layers for large text classification datasets. Furthermore, Santos and Gatti [2] presented a CNN architecture that jointly used character-level, word-level and sentence-level representations for sentiment analysis.

Gehring et al. [5] have introduced a fully CNN architecture for sequence-to-sequence learning in machine translation and their architecture outperformed strong recurrent models on benchmark datasets. Also, Meng et al. [4] proposed gated convolutional neural networks for statistical machine translation and achieved state-of-the-art performance on two NIST Chinese-English translation tasks. Yin et al. [10] used a CNN model for question answering and got an outstanding performance. They believe that CNN's have high performance in question answering task. Dauphin et al. [7] proposed a gating mechanism to control which information flows in the network. Their proposed gated CNN achieved the state-of-the-art results on WiKiText-103.

CNN's also have been already successfully applied to a variety of computer vision and sound recognition problems. Dense convolutional network (DenseNet) [18] is one of the latest neural networks for visual object recognition and it obtains significant improvements over benchmark datasets. In their proposed architecture, each layer connects to every other layer in a feed-forward fashion. Bai et al. [45] have proposed a novel convolutional neural network variant (named MSP-Net) which can classify images that contain text or not. The MSP-Net was tested on several datasets and demonstrated acceptable accuracy. Li et al. [46] presented a robust tracking algorithm based on CNN for learning feature representations of the target object. The CNN-based deep track achieved the comparable tracking speed. Andrearczyk and Whelan [47] developed a new method for the analysis of dynamic texture by using CNN's and obtained high accuracy in all tested datasets. Babaee et al. [48] presented a new background subtraction method which has used CNN to perform the segmentation task. Abdel-Hamid et al. [12] used CNN's for speech recognition and obtained lower error rates. CNN's also are used for text clustering. For instance, Xu et al. [49] proposed a convolutional neural network framework for short text clustering, which can incorporate more semantic features of the text. They tested their framework on three datasets and enhanced the performance of text clustering.

Recent studies have demonstrated that CNN's gives better performance compared to other deep learning methods. Lee et al. [3] proposed a weakly supervised learning method based on a CNN for sentiment classification and it was 11 times faster than RNN-based attention model. One of the problems of RNN's is that the former words have less effect on final representation but CNN's are not [50]. Another major drawback of RNN is high time complexity [49]. Visin et al. [51] tried to replace the convolutional layer with four RNN's for object recognition, but their model did not outperform state-of-the-art CNN's on all tested benchmark datasets. Xu et al. [49] believe that CNN is better to learn non-biased implicit features than LSTM and gated recurrent unit (GRU). Also, unlike CNN's, LSTM's and Multi-Layer Perceptrons (MLP's) are memory bandwidth limited [52].

### 2.3. Highway networks

Highway network was introduced to overcome the very deep neural network limitations, such as training, vanishing and exploding gradient. The highway models improve the performance of both deep and shallow neural networks [23]. Zilly et al. [23] have proposed Recurrent Highway Networks(RHN) which is the combination of highway and RNN. The RHN uses multiple layers of highway networks for learning very deep recurrent neural networks. Also, Kim et al. [53] employed CNN, LSTM and a highway network for neural language modeling. They applied the highway network's output as the input to the LSTM and achieved state-of-art results. Similarly, Zhang et al. [54] presented a highway LSTM network for speech recognition task and their architecture outperformed all previous works.

## 3. Proposed architecture and method

In this section, we present in detail our proposed model architecture and improved word vectors as the inputs of the architecture.

### 3.1. Improved Word Vectors (IWV)

The IWV is based on combining of Part-of-Speech (POS) tagging techniques, lexicon-based approaches and Word2Vec methods. The main architecture of the IWV has been shown in figure 1.

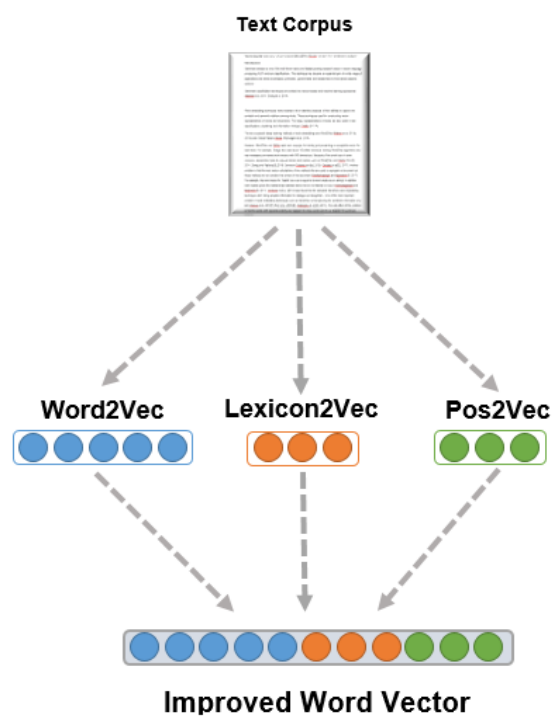

Figure1: The architecture of the Improved Word Vector (IWV)

### 3.1.1. Lexicon2Vec (L2V)

The sentiment and emotion lexicons are lists of phrases and words which have polarity scores and can be used to analyze texts. Each lexicon contains words and their values which are the sentiment scores for those words. There are various sentiment and emotion lexicons that can be used, but choosing the proper combination of lexicons is very important because some combinations of lexicons are more accurate than others. We selected seven lexicons as our resources and assigned vectors to each word. Selected lexicons are as follows:

- NRC Emoticon Lexicon [55,56,57]
- National Research Council Canada (NRC) Emoticon Affirmative Context Lexicon and NRC Emoticon Negated Context Lexicon [55,56,57]
- NRC Hashtag Sentiment Lexicon [55,56,57]
- NRC Hashtag Affirmative Context Sentiment Lexicon and NRC Hashtag Negated Context Sentiment Lexicon [55,56,57]
- Amazon Laptop Sentiment Lexicon [58]
- SemEval-2015 English Twitter Sentiment Lexicon [59,56]
- Yelp Restaurant Sentiment Lexicon [58]

### 3.1.2. POS2Vec (P2V)

Part-of-speech (POS) tagging is an important and fundamental step in Natural Language Processing which is the process of assigning to each word of a text the proper POS tag. The Part-of-speech gives a large amount of information about a word and its neighbors, syntactic categories of words (nouns, verbs, adjectives, adverbs, etc.) and similarities and dissimilarities between them. We converted each generated POS tag to a constant vector and concatenated with Word2Vec/GloVe vectors. As a result, Word2Vec/GloVe vectors will have syntactic information of words.

### 3.1.3. Word2Vec and GloVe

Word2Vec is based on continuous Bag-of-Words (CBOW) and Skip-gram architectures which can provide high-quality word embedding vectors. CBOW predicts a word given its context and Skip-gram can predict the context given a word. The generated vectors of words which appear in common contexts in the corpus are located close to each other in the vector space. GloVe word embedding is a global log-bilinear regression model and is based on co-occurrence and factorization of a matrix in order to get vectors.

### 3.2. Proposed architecture

The proposed model architecture is shown in figure 2 and it contains some blocks with different filter region sizes. First, each filter performs a convolution on the IWV and generates a feature map. Then, each generated feature map is used as an input of a block. Next, we apply a max pooling over the feature maps. The last step, a softmax layer receives the feature vectors as input and uses it to classify sentences. In the following, more details are provided about the model.

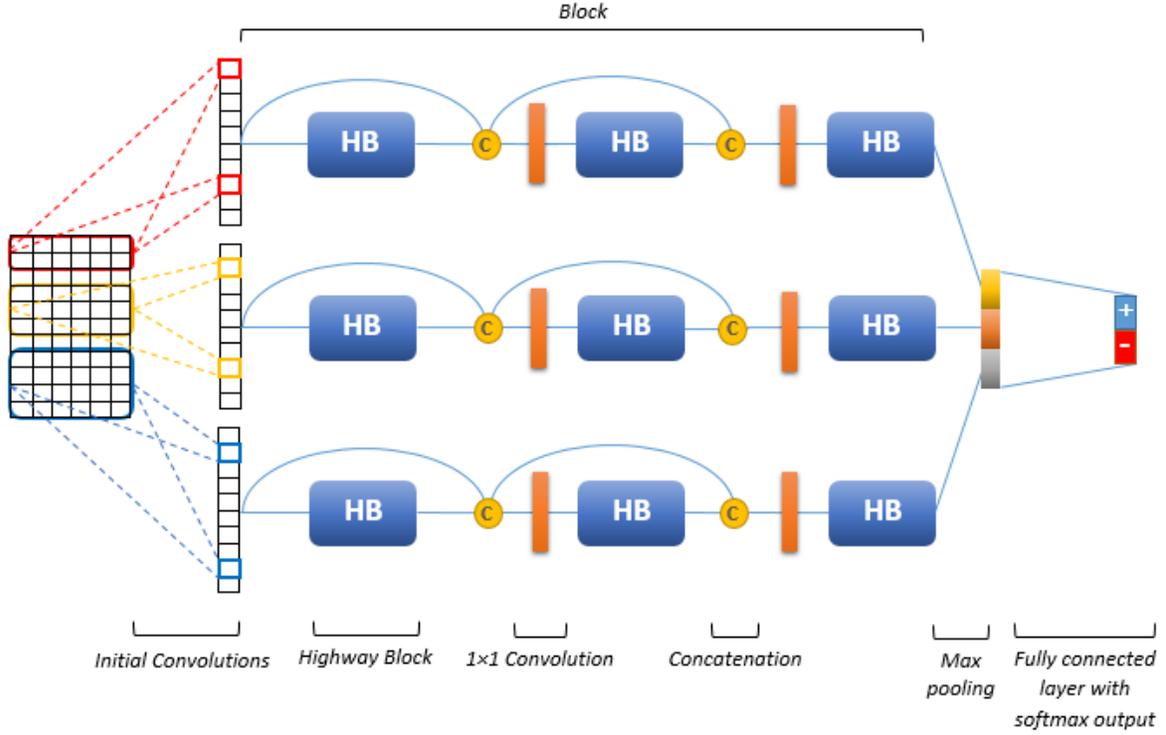

Figure 2: The proposed model architecture

### 3.2.1. Initial convolutional layer

Each word in a sentence of length $n$ is associated with an $M$-dimensional vector. Let $x_i \in R^M$ be the $M$-dimensional improved word vector corresponding to the $i$-th word in the sentence. Assume $X_{i:i+j}$ denote the concatenation of improved word vectors $x_i$ to $x_{i+j}$. A convolution operation involves a filter $W_c \in R^{h*M}$, applied to a window of $h$ words to produce a new feature. A feature map $c_i$ is defined as below:

$$c_i = ReLU\,(\,BN\,(W_c\,.\,X_{i:i+h-1} + b\,))  \tag{1}$$

Here $BN$ is batch normalization transform, $ReLU$ (Rectified Linear Unit) is a non-linear activation function, $b$ is a bias term and (.) denotes the convolutional operator.

### 3.2.2 Batch normalization

Batch normalization introduced by Ioffe and Szegedy [37] and it has some advantages such as reduced internal covariate shift, improves gradient flow through the network, allows higher learning rates and network trains faster. Assume the layer that needs to be normalized has a $d$ dimensional input $Z = (z_1, z_2, z_3, \ldots, z_d)$. First, we will normalize the *k*th dimension as follows:

$$\hat{z}_k = \frac{(z_k - \mu_\beta)}{\sqrt{\delta_\beta^2 + \epsilon}} \tag{2}$$

where $\delta_\beta^2$ and $\mu_\beta$ are the variance and mean of mini-batch respectively, and $\epsilon$ is a constant value. For increasing the ability of the representation, the normalized input $\hat{Z}_k$ is further transformed into:

$$y_k = BN_{\gamma,\beta}(z_k) = \gamma \hat{z}_k + \beta \tag{3}$$

where $\beta$ and $\gamma$ are learned parameters.

### 3.2.3. Highway block

Highway Network has proposed by Srivastava et al. [22]. The proposed highway block contains convolution layers, batch normalization and activation functions. We found that the batch normalization will improve the accuracy of the highway block. Let $c$ is the input to the highway block and $y$ is the highway block's output. The highway block does the following:

$$y = t \odot g\left(BN(c.W_H + b_H)\right) + (1-t) \odot c \tag{4}$$

$$t = h\left(BN(c.W_T + b_T)\right) \tag{5}$$

Where $BN$ is batch normalization transform, $t$ is called the transform gate, $(1-t)$ is called the carry gate, $W_T$ and $W_H$ are square matrices, $b_H$ and $b_T$ are bias terms, $g$ and $h$ are non-linear transformation functions and the dot operator (⊙) is used to denote element-wise multiplication. Here $g$ and $h$ are $ReLU$ and $Sigmoid$ function, respectively.

### 3.2.4. Bottleneck layer

He et al. [36] and Szegedy et al. [38] have noted that a 1×1 convolution layer that can be used to obtain a reduced dimensionality of the input feature maps, and thus to improve computational efficiency. We found that the bottleneck layer is effective for the proposed architecture, so after each concatenation, we use bottleneck layers to reduce the dimensionality of the input of highway blocks. In our model, the output dimension of the 1×1 convolution is 100.

### 3.2.5. Max-over-time pooling

The outputs of the last highway blocks are then passed to the max pooling layer. Let $c = [c_1, c_2, ..., c_{n-h+1}]$ be a feature map of the sentence $\{X_{1:h}, X_{2:h+1}, ..., X_{n-h+1}\}$. By applying a max-over-time pooling operation over the feature map we will take the maximum value $\hat{c} = Max\{c\}$ as the feature corresponding to this filter. This layer can extract the most important features.

### 3.2.6. Softmax layer

The model is topped by a softmax classifier layer that predicts the probability distribution over classes. The output of the max pooling layer is passed to a fully connected softmax layer. Assume, $q$ is a vector of the inputs to the output layer and there are *k* output labels. The softmax function is defined as follows:

$$softmax(q) = \frac{\exp(q)}{\sum_{j=1}^{k} \exp(q_j)} \quad (6)$$

## 4. Experiments

In this section, we study the empirical performance of the MBCH on benchmark datasets and then compare it to other models. In addition, we describe the datasets and experimental evaluations to show the effectiveness of the MBCH.

### 4.1. Datasets

Summary statistics of the datasets used in our study are listed in table1. More details are described as follows:

**CR:** Customer reviews of 14 products obtained from Amazon and classified into positive and negative reviews [60].

**Subj:** Subjectivity data set where the goal is to classify a sentence as being subjective or objective [61].

**Yelp:** It contains 1000 randomly sampled reviews for restaurants from Yelp' 13 dataset, which was introduced by Kotzias et al. [42].

**MPQA:** Opinion polarity subtask of the MPQA dataset [62].

**IMDB:** It contains 1000 randomly sampled movie reviews from IMDB dataset, which was introduced by Kotzias et al. [42].

Table1: Statistical information of the datasets. N: Dataset size. |V|: Vocabulary size. Positive: Number of positive examples. Negative: Number of negative examples

| Dataset | Tasks | N | |V| | Positive | Negative |
|---|---|---|---|---|---|
| CR | sentiment (products reviews) | 3803 | 5572 | 2397 | 1406 |
| Subj | subjectivity/objectivity | 10000 | 21317 | 5000 | 5000 |
| Yelp | sentiment (restaurants reviews) | 1000 | 2042 | 500 | 500 |
| MPQA | opinion polarity | 10604 | 6234 | 3311 | 7293 |
| IMDB | sentiment (movies reviews) | 1000 | 3076 | 500 | 500 |

### 4.2 Implementation details

Our implementations of MBCH and IWV were GPU-based - training the framework on four GeForce GTX Titan X GPUs. Tensorflow was used for implementing and training the model in our research. We used 10-fold cross-validation (CV) to evaluate the accuracy of the model. We use filter sizes of (2,3,4,5) and (2,3,4,5,6,7) with 500 feature maps each, mini-batch size of 16 and learning rate 3e-4. Dropout was not used in our research. Baseline configuration is shown in table 2.

Table2: Baseline configurations

| Models | Input word vector | Filter region size | N. of Filters | Optimizer | $l_2$ norm constraint | Learning rate | Batch size |
|---|---|---|---|---|---|---|---|
| **MBCH-4F** | IWV | (2,3,4,5) | 500 each | Adam | 0.2 | 3e-4 | 16 |
| **MBCH-6F** | IWV | (2,3,4,5,6,7) | 500 each | Adam | 0.2 | 3e-4 | 16 |

### 4.3. Results

We empirically demonstrate the proposed model's effectiveness on benchmark datasets and compare its results to other models. As stated above, 7 sentiment lexicons were used to extract and generate the lexicon vectors. We only used unigram scores in our research. Sentiment scores of each word are extracted from all lexicons and are normalized. If a word doesn't exist in any lexicons, its score will be zero. The statistics of the lexicons are given in Table 3.

Table 3: Statistics of the lexicons which were used in the research

| Lexicon | Positive | Negative | Neutral | Total | Scores Ranges |
|---|---|---|---|---|---|
| SemEval-2015 English Twitter Sentiment Lexicon | 776 | 726 | 13 | 1515 | -0.984 to +0.984 |
| NRC Emoticon Lexicon | 38312 | 24156 | 0 | 62468 | -4.999 to +5.0 |
| NRC Hashtag Sentiment Lexicon | 32048 | 22081 | 0 | 54129 | -6.925 to +7.526 |
| Amazon Laptop Sentiment Lexicon | 14651 | 11926 | 0 | 26577 | -5.27 to +3.702 |
| NRC Emoticon Affirmative Context Lexicon and NRC Emoticon Negated Context Lexicon | 28025 | 27121 | 0 | 55146 | -5.844 to +4.495 |
| Yelp Restaurant Sentiment Lexicon | 20347 | 18927 | 0 | 39274 | -4.44 to +3.798 |
| NRC Hashtag Affirmative Context Sentiment Lexicon and NRC Hashtag Negated Context Sentiment Lexicon | 19502 | 24447 | 0 | 43949 | -10.025 to +10.661 |

We compared the performance of our model with the state-of-the-art models listed below on the five benchmark datasets.

**Tree-CRF:** It was proposed by Nakagawa et al. [63] for sentiment analysis and it is a dependency tree based method using CRF with hidden variables.

**NBSVM** and **MNB:** Naive Bayes support vector machines and multinomial naive Bayes with uni-bigrams proposed by Wang and Manning [64]

**F-Dropout** and **G-Dropout**: Fast dropout and Gaussian dropout from Wang and Manning [65]

**BiLSTM-CRF + CNN:** A combination of three methods, BiLSTM, CRF and CNN for sentiment analysis [66]

**CNN-rand**: Convolutional neural network using random initialization vectors [1]

**CNN-non-static:** Convolutional neural network using pre-trained word2vec vectors for each task [1]

**CNN-multichannel:** Convolutional neural network using multichannel architecture [1]

**CNN-non-static (W2V+GloVe):** Convolutional neural network using combined pre-trained Word2Vec and GloVe vectors for each task [29].

**Logistic w/ BOW on Documents**: Logistic regression classifier on a bag of words representation at the document level [42].

**Logistic w/ BOW on Sentences:** Logistic regression classifier on a bag of words representation at the sentence level [42].

**Logistic w/ Embeddings on Documents:** logistic regression classifier using embedding vectors which are provided by the ConvNet [42].

**GICF w/ Embeddings on Sentences**: Group-instance cost function using embedding vectors [42].

One of the strengths of MBCH is that it is relatively shallow compared to other Deep Neural Networks in this area [17,15], while simultaneously achieving high accuracy on the benchmark datasets. The results of our models against other models are presented in table 4.

Table 4: The results of our models (MBCH-4F & MBCH-6F) against other models. The best results are highlighted in boldface

| Model | Subj | Yelp | CR | MPQA | IMDB |
|---|---|---|---|---|---|
| CNN-rand | 89.6 | - | 79.8 | 83.4 | - |
| CNN-non-static | 93.4 | - | 84.3 | 89.5 | - |
| CNN-multichannel | 93.2 | - | 85.0 | 89.4 | - |
| CNN-non-static (W2V+GloVe) | 93.6 | | 84.6 | 89.5 | |
| Logistic w/ BOW on Documents (L-BD) | - | 91.2 | - | - | 86.2 |
| Logistic w/ BOW on Sentences (L-BS) | - | 78.1 | - | - | 81.8 |
| MNB | 93.6 | - | 80.0 | 86.3 | - |
| BiLSTM-CRF + CNN | - | - | 85.4 | - | - |
| Logistic w/ Embeddings on Documents (L-ED) | - | 81.0 | - | - | 58.2 |
| GICF w/ Embeddings on Sentences (GICF-ES) | - | 88.7 | - | - | 88.5 |
| G-Dropout | 93.4 | - | 82.1 | 86.1 | - |
| F-Dropout | 93.6 | | 81.9 | 86.3 | |
| NBSVM | 93.2 | - | 81.8 | 86.3 | - |
| Tree-CRF | - | - | 81.4 | 86.1 | - |
| **MBCH-4F** | 94.0 | **93.6** | 85.5 | **90.5** | **92.0** |
| **MBCH-6F** | **94.1** | 93.5 | **85.7** | 90.4 | **92.0** |

Table 4 shows the experimental results on five datasets: Subj, Yelp, CR, MPQA and IMDB. It can be observed that MBCH models outperform all the other approaches in all the five datasets. The MBCH-4F gives a substantial improvement of 3.5% compared to GICF-ES on the IMDB dataset. In addition, it gives an improvement of 2.4% compared to LBD on the Yelp dataset. Also, the MBCH-4F obtained the best accuracy (90.5%) on the MPQA dataset, 1% higher than the CNN-non-static model. As seen from the results, the MBCH-6F achieved the highest accuracy of 94.1% on the Subj dataset. Also, it obtained an accuracy rate of 92%, on the IMDB dataset. For the CR dataset, the MBCH-6F achieved the highest accuracy (85.7%) in comparison with the other models.

### 4.3.1 Effect of filter size

In this study, we also explored the effect of combining different filter sizes on the five selected datasets and used 10-fold cross-validation to evaluate the accuracy. Table 5 shows the different combinations of filters which were used in this research.

Table 5: Different combinations of filters

| Name | Filters |
|------|---------|
| A | (2,3,4) |
| B | (3,4,5) |
| C | (4,5,6) |
| D | (5,6,7) |
| E | (2,3,4,5) |
| F | (3,4,5,6) |
| G | (4,5,6,7) |
| H | (2,3,4,5,6,7) |

According to figure 3, it can be seen that (4,5,6) and (4,5,6,7) perform worst on the CR, IMDB, MPQA and Subj. However, for the Yelp dataset, filter sizes of (5,6,7) and (3,4,5,6) have lower accuracy than others. As seen from the results that (2,3,4), (2,3,4,5) and (2,3,4,5,6,7) perform better than all other filter sizes.

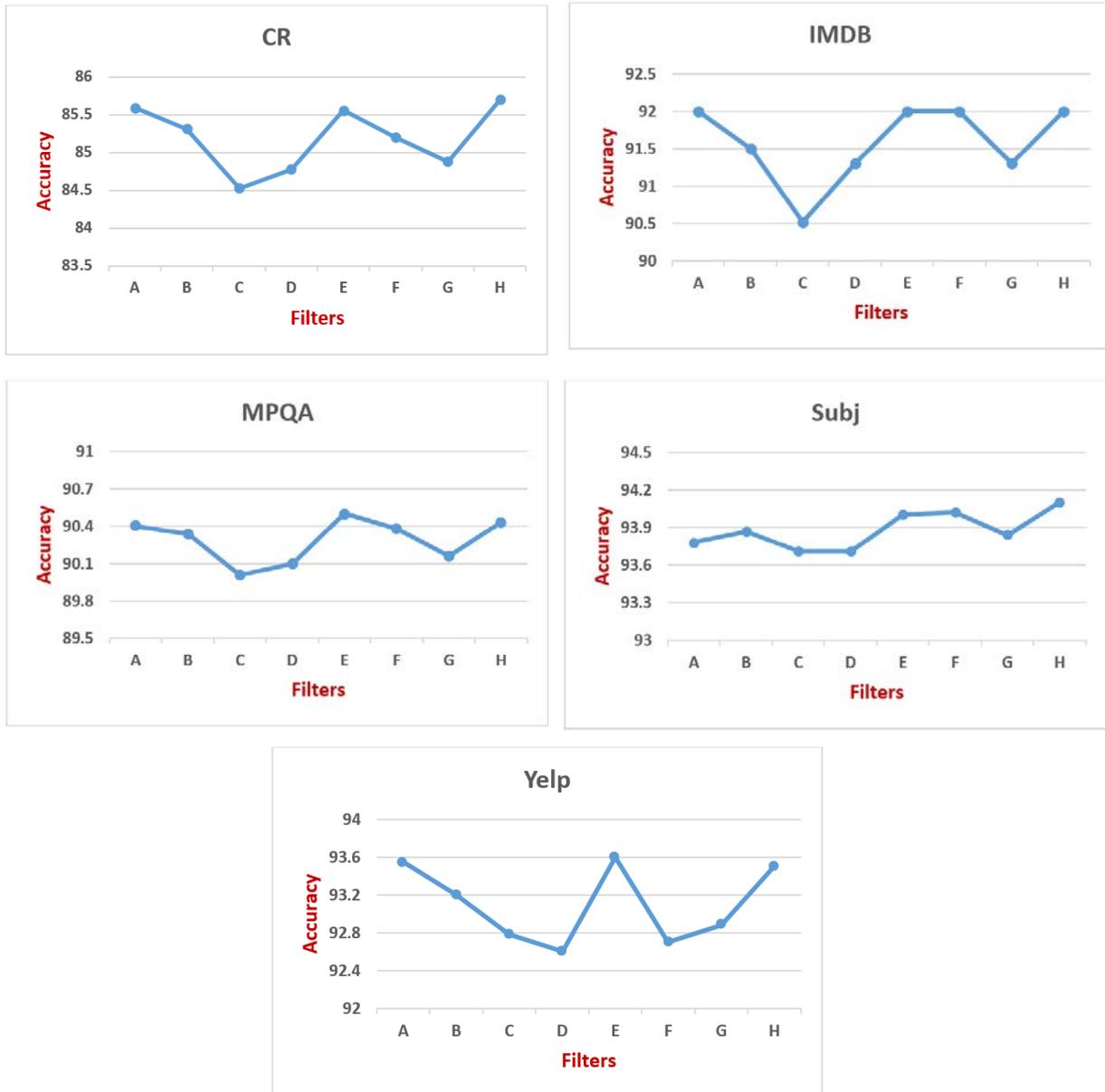

Figure3: Effect of filter size with several sizes on the five benchmark datasets

### 4.3.2 Effect of number of feature maps

In this section, we investigate the effect of the number of feature maps on the benchmark datasets. We consider (2,3,4,5) and the number of feature maps of 100, 200, 300, 400 and 500. Also, the 10-fold cross-validation is used to evaluate the accuracy. The results can be seen in figure 4.

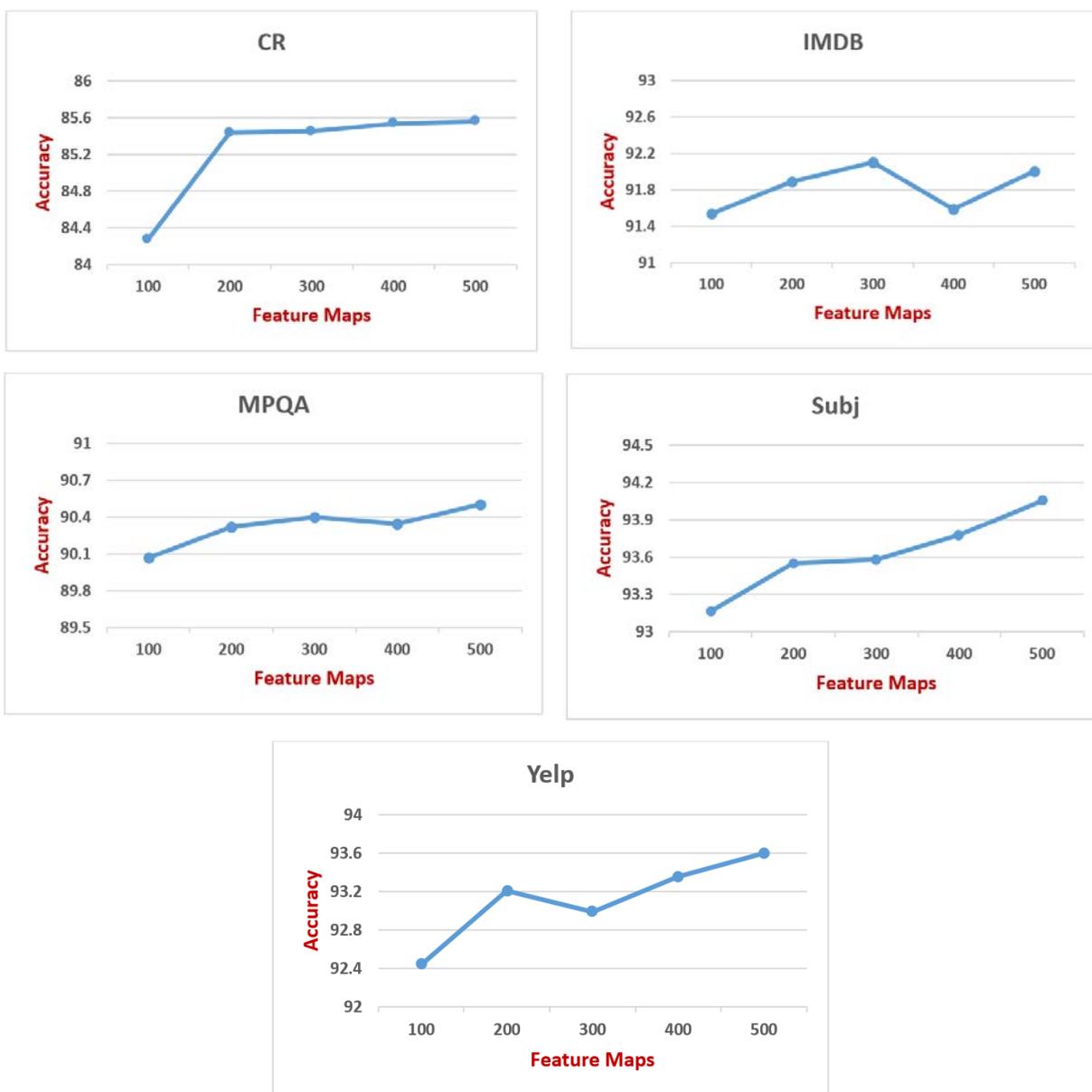

Figure4: Effect of the number of feature maps on the five benchmark datasets

As can be seen in figure 4, feature maps of 500 have the highest accuracy among all the datasets. However, in the IMDB dataset, accuracies of feature maps of 300 and 500 are approximately equal. In the Yelp dataset, the accuracy of the feature map of 200 is higher than the feature map of 300, but it is reversed on other datasets. The accuracy of the feature maps of 400 is lower than the feature maps of 300 in the MPQA and IMDB datasets.

## 5. Conclusion

In this paper, we proposed a novel architecture for text classification tasks of Sentiment Analysis, Subjectivity/Objectivity Analysis, and Opinion Polarity and achieved the state-of-art results on different benchmark datasets. The MBCH is based on recent advances in convolutional neural networks while being relatively shallow and improving the accuracy of text categorization. Significantly, the accuracy of the IMDB and Yelp increased by 3.5% and 2.4% respectively. Also, we proposed a new method - IWV - to improve the accuracy of well-known pre-trained word embeddings which are used as inputs of CNN's. The IWV was based on the combination of four approaches such as lexicon-based approaches, POS tagging techniques and Word2Vec methods. According to the results, MBCH with IWV are very effective and useful for text classification tasks.

**References:**

[1] Kim, Y., Convolutional Neural Networks for Sentence Classification. Proceedings of the 2014 Conference on Empirical Methods in Natural Language Processing (EMNLP 2014), 2014.

[2] D. Santos, M. Gatti, Deep Convolutional Neural Networks for Sentiment Analysis of Short Texts, in COLING 2014, 25th International Conference on Computational Linguistics, 2014, pp. 69-78.

[3] G. Lee, J. Jeong, S. Seo, C. Kim, P. Kang, Sentiment classification with word localization based on weakly supervised learning with a convolutional neural network. Know.-Based Syst., **152**(C), 2018, pp. 70-82.

[4] F. Meng, Z. Lu, M. Wang, H. Li, W. Jiang, Q. Liu, Encoding Source Language with Convolutional Neural Network forMachine Translation. Proceedings of the 53rd Annual Meeting of the Association for Computational Linguistics and the 7th International Joint Conference on Natural Language Processing, 2015, pp. 20-30.

[5] J. Gehring, M. Auli, D. Grangier, Y. Dauphin, A Convolutional Encoder Model for Neural Machine Translation. Proceedings of the 55th Annual Meeting of the Association for Computational Linguistics, 1, 2017, pp. 123–135.

[6] N. Kalchbrenner, E. Grefenstette, B. Phil, A Convolutional Neural Network for Modelling Sentences. Proceedings of the Conference 52nd Annual Meeting of the Association for Computational Linguistics, ACL, 1, 2014, pp. 655–665.

[7] Y. N. Dauphin, A. Fan, M. Auli, D. Grangier, Language Modeling with Gated Convolutional Networks. Proceedings of the International Conference on Machine Learning (ICML), 2017, pp. 933-941

[8] X. Qiu, X. Huang, Convolutional neural tensor network architecture for community-based question answering, in Proceedings of the 24th International Conference on Artificial Intelligence, AAAI Press: Buenos Aires, Argentina, 2015, pp. 1305-1311.

[9] L. Dong, F. Wei, M. Zhou, K. Xu, Question Answering over Freebase with Multi-Column Convolutional Neural Networks, Proceedings of the 53rd Annual Meeting of the Association for Computational Linguistics and the 7th International Joint Conference on Natural Language Processing, 2015, pp. 260-269.

[10] W. Yin, M.Y., B. Xiang, B. Zhou, H. Schutze, Simple Question Answering by Attentive Convolutional Neural Network. Proceedings of COLING 2016, the 26th International Conference on Computational Linguistics: Technical Papers 2016, pp. 1746–1756.

[11] R. Collobert, J. Weston, L. Bottou, M. Karlen, K. Kavukcuoglu,P. Kuksa, Natural Language Processing (Almost) from Scratch, J. Mach. Learn. Res., 12, 2011, pp. 2493-2537.

[12] O. Abdel-Hamid, A.r. Mohamed, H. Jiang, G. Penn, Applying Convolutional Neural Networks concepts to hybrid NN-HMM model for speech recognition, IEEE International Conference on Acoustics, Speech and Signal Processing (ICASSP), 2012, pp. 4277-4280.

[13] O. Abdel-Hamid, A.r. Mohamed, H. Jiang, L. Deng, G. Penn, D. Yu, Convolutional Neural Networks for Speech Recognition, IEEE/ACM Transactions on Audio, Speech, and Language Processing, 22, 2014, pp. 1533-1545.

[14] J. Gu, Z. Wang, J. Kuen, L. Ma, A. Shahroudy, B. Shuai, T. Liu, X. Wang, G. Wang, J. Cai, T. Chen Recent Advances in Convolutional Neural Networks, Pattern Recogn., 77, 2018, pp. 354-377.

[15] T. Sercu , V. Goel, Advances in Very Deep Convolutional Neural Networks for LVCSR. Proceedings of the International Speech Communication Association (INTERSPEECH), 2016, pp. 3429–3433.

[16] S.N.B. Bhushan, A. Danti, Classification of text documents based on score level fusion approach, Pattern Recogn. Lett., 94, 2017, pp. 118-126.

[17] A. Conneau, S. Holger, L. Barrault, Y. Lecun, Very Deep Convolutional Networks for Text Classification, Proceedings of the 15th Conference of the European Chapter of the Association for Computational Linguistics, 1, 2017, pp. 1107-1116.

[18] G. Huang, Z. Liu, L. v. d. Maaten, K. Q. Weinberger, Densely Connected Convolutional Networks, IEEE Conference on Computer Vision and Pattern Recognition (CVPR), 2017, pp. 2261-2269.

[19] Y. Zhang, W.C., N. Jaitly, Very deep convolutional networks for end-to-end speech recognition. IEEE International Conference on Acoustics, Speech and Signal Processing (ICASSP), 2017, pp. 4845-4849.

[20] S. Hochreiter, Y. Bengio, P. Frasconi, J. Schmidhuber, Gradient Flow in Recurrent Nets: the Difficulty of Learning Long-Term Dependencies, A Field Guide to Dynamical Recurrent Neural Networks- IEEE Press, 2001.

[21] Y. Bengio, P. Simard, P. Frasconi, Learning long-term dependencies with gradient descent is difficult, IEEE Transactions on Neural Networks, **5**(2), 1994, pp. 157-166.

[22] R. K. Srivastava, K. Greff, J. Schmidhuber, Training Very Deep Networks. Proceedings of the 28th International Conference on Neural Information Processing Systems, 2, 2015, pp. 2377-2385.

[23] J.G. Zilly, R. K. Srivastava, J. Koutník, J. Schmidhuber, Recurrent Highway Networks, arXiv preprint arXiv:1607.03474, 2016.

[24] T. Mikolov, I. Sutskever, K. Chen, G. Corrado, J. Dean, Distributed representations of words and phrases and their compositionality, Proceedings of the 26th International Conference on Neural Information Processing Systems, 2, 2013, pp. 3111-3119.

[25] T. Mikolov, K. Chen, G. Corrado, J. Dean, Efficient Estimation of Word Representations in Vector Space. CoRR, 2013, abs/1301.3781.

[26] J. Pennington, R. Socher, C. Manning, Glove: Global Vectors for Word Representation. Proceedings of the 2014 Conference on Empirical Methods in Natural Language Processing (EMNLP), 2014, pp. 532–1543.

[27] M. Giatsoglou, M. Vozalis, K. Diamantaras, A. Vakali, G. Sarigiannidis, K. Chatzisavvas, Sentiment analysis leveraging emotions and word embeddings. Expert Systems with Applications, 69, 2017, pp. 214-224.

[28] O. Araque, I. Corcuera-Platas, J. F. Snchez-Rada, C. Iglesias, Enhancing deep learning sentiment analysis with ensemble techniques in social applications, Expert Syst. Appl., 77, 2017, pp. 236-246.

[29] Y. Zhang, B. Wallace, A Sensitivity Analysis of (and Practitioners' Guide to) Convolutional Neural Networks for Sentence Classification, arXiv, 1510.03820v4., 2015.

[30] J. Camacho-Collados, M. Pilevar, R. Navigli, NASARI: Integrating explicit knowledge and corpus statistics for a multilingual representation of concepts and entities, Artificial Intelligence, 240, 2016.

[31] A. Caliskan, J. Bryson, A. Narayanan, Semantics derived automatically from language corpora contain human-like biases. Science, 356, 2017, pp. 183 -186.

[32] Y. Wang, M. Huang, X. Zhu, L. Zhao, Attention-based LSTM for Aspect-level Sentiment Classification, Proceedings of the 2016 Conference on Empirical Methods in Natural Language Processing, 2016, pp. 606-615.

[33] M. Iyyer, V. Manjunatha, J. Boyd-Graber, H. Daumé, Deep Unordered Composition Rivals Syntactic Methods for Text Classification. Proceedings of the 53rd Annual Meeting of the Association for Computational Linguistics and the 7th International Joint Conference on Natural Language Processing, 1, 2015, pp. 1681-1691.

[34] Y. Ren, R. Wang, D. Ji, A topic-enhanced word embedding for Twitter sentiment classification, Inf. Sci., 369, 2016, pp. 188-198.

[35] M. Kamkarhaghighi, M. Makrehchi, Content Tree Word Embedding for document representation. Expert Systems with Applications, 90, 2017, pp. 241-249.

[36] K. He, X. Zhang, S. Ren, J. Sun, Deep Residual Learning for Image Recognition, IEEE Conference on Computer Vision and Pattern Recognition (CVPR), 2016, pp. 770-778.

[37] S. Ioffe, C. Szegedy, Batch Normalization: Accelerating Deep Network Training by Reducing Internal Covariate Shift, Proceedings of the 32Nd International Conference on International Conference on Machine Learning (ICML'15), 37, 2015, pp. 448-456.

[38] C. Szegedy, V. Vanhoucke, S. Ioffe, J. Shlens, Z. Wojna, Rethinking the Inception Architecture for Computer Vision. IEEE Conference on Computer Vision and Pattern Recognition (CVPR), 2016, pp. 2818-2826.

[39] A. Severyn, A. Moschitti, Twitter Sentiment Analysis with Deep Convolutional Neural Networks, Proceedings of the 38th International ACM SIGIR Conference on Research and Development in Information Retrieval, 2015, pp. 959-962.

[40] P. Qin, W. Xu, J. Guo, An Empirical Convolutional Neural Network Approach for Semantic Relation Classification, Neurocomput., 190, 2016, pp. 1-9.

[41] D. Zeng, K. Liu, S. Lai, J. Zhao, Relation classification via convolutional deep neural network, the 25th International Conference on Computational Linguistics: Technical Papers, 2014, pp. 2335-2344.

[42] D. Kotzias, M. Denil, N. Freitas, P. Smyth From Group to Individual Labels Using Deep Features, Proceedings of the 21th ACM SIGKDD International Conference on Knowledge Discovery and Data Mining, 2015, pp. 597-606.

[43] D. Tang, B. Qin, T. Liu, Learning Semantic Representations of Users and Products for Document Level Sentiment Classification, ACL, 2015.

[44] X. Zhang, J. Zhao, Y. LeCun, Character-level convolutional networks for text classification, Proceedings of the 28th International Conference on Neural Information Processing Systems, 1, 2015, pp. 649-657.

[45] X. Bai, B. Shi, C. Zhang, X. Cai, L. Qi, Text/Non-text Image Classification in the Wild with Convolutional Neural Networks, Pattern Recogn., 66, 2017, pp. 437-446.

[46] H. Li, Y. Li, F. Porikli, DeepTrack: Learning Discriminative Feature Representations Online for Robust Visual Tracking, IEEE Transactions on Image Processing, 25(4), 2016.

[47] V. Andrearczyk, P.W., Convolutional neural network on three orthogonal planes for dynamic texture classification, Pattern Recogn., 2018, 76, pp. 36-49.

[48] M. Babaee, D. Dinh, G. Rigoll, A deep convolutional neural network for video sequence background subtraction, Pattern Recogn., 76, 2018, pp. 635-649.

[49] J. Xu, B. Xu, P. Wang, S. Zheng, G. Tian, J. Zhao, Self-Taught Convolutional Neural Networks for Short Text Clustering, Neural Networks, 88, 2016, pp. 22-31.

[50] Y. Li, B. Wei, Y. Liu, L. Yao, H. Chen, F. Yu, W. Zhu, Incorporating Knowledge into neural network for text representation, Expert Systems with Applications, 96, 2017, pp. 103-114.

[51] F. Visin, K. Kastner, K. Cho, M. Matteucci, A. Courville, Y. Bengio, ReNet: A Recurrent Neural Network Based Alternative to Convolutional Networks, arXiv preprint arXiv:1505.00393, 2015.

[52] N. P. Jouppi, C. Young, N. Patil, D. Patterson, G. Agrawal, R. Bajwa, S. Bates, S. Bhatia, N. Boden, A. Borchers, R. Boyle, P. Cantin, C. Chao, C. Clark, J. Coriell, M. Daley, M. Dau, J. Dean, B. Gelb, T. Ghaemmaghami, R. Gottipati, W. Gulland, R. Hagmann, C. Richard Ho, D. Hogberg, J. Hu, R. Hundt, D. Hurt, J. Ibarz, A. Jaffey, A. Jaworski, A. Kaplan, H. Khaitan, D. Killebrew, A. Koch, N. Kumar, S. Lacy, J. Laudon, J. Law, D. Le, C. Leary, Z. Liu, K. Lucke, A. Lundin, G. MacKean, A. Maggiore, M. Mahony, K. Miller, R. Nagarajan, R. Narayanaswami, R. Ni, K. Nix, T. Norrie, M. Omernick, N. Penukonda, A. Phelps, J. Ross, M. Ross, A. Salek, E. Samadiani, C. Severn, G. Sizikov, M. Snelham, J. Souter, D. Steinberg, A. Swing, M. Tan, G. Thorson, B. Tian, H. Toma, E. Tuttle, V. Vasudevan, R. Walter, W. Wang, E. Wilcox, D. Yoon, In-Datacenter Performance Analysis of a Tensor Processing Unit, SIGARCH Comput. Archit. News, 45(2), 2017, pp. 1-12.

[53] Y. Kim, Y. Jernite, D. Sontag, A. M. Rush, Character-aware Neural Language Models, Proceedings of the Thirtieth AAAI Conference on Artificial Intelligence, 2016, pp. 2741-2749.

[54] Y. Zhang, G.Chen, D. Yu, K. Yaco, S. Khudanpur, J. Glass, Highway long short-term memory RNNS for distant speech recognition. IEEE International Conference on Acoustics, Speech and Signal Processing (ICASSP), 2016, pp. 5755-5759.

[55] S. Mohammad, K. Svetlana, X. Zhu, NRC-Canada: Building the State-of-the-Art in Sentiment Analysis of Tweets, In Proceedings of the seventh international workshop on Semantic Evaluation Exercises (SemEval-2013), 2013, pp. 321-327.

[56] S. Kiritchenko, X. Zhu, S. Mohammad, Sentiment Analysis of Short Informal Texts, Journal of Artificial Intelligence Research (JAIR), 50, 2014, pp. 723-762.

[57] X. Zhu, S. Kiritchenko, S. Mohammad, Recent Improvements in Sentiment Analysis of Tweets, In Proceedings of the eigth international workshop on Semantic Evaluation Exercises (SemEval-2014), 2014.

[58] S. Kiritchenko, X. Zhu, C. Cherry, S. Mohammad, Detecting Aspects and Sentiment in Customer Reviews, In Proceedings of the eigth international workshop on Semantic Evaluation Exercises (SemEval-2014), 2014.

[59] S. Rosenthal, P. Nakov, S. Kiritchenko, S. Mohammad, A. Ritter, V. Stoyanov, Sentiment Analysis in Twitter, In Proceedings of the ninth international workshop on Semantic Evaluation Exercises (SemEval-2015), 2015.

[60] M. Hu, B. Liu, Mining and Summarizing Customer Reviews, Proceedings of the Tenth ACM SIGKDD International Conference on Knowledge Discovery and Data Mining, 2004, pp. 168-177.

[61] B. Pang, L. Lee, Seeing stars: exploiting class relationships for sentiment categorization with respect to rating scales, Proceedings of the 43rd Annual Meeting on Association for Computational Linguistics, 2005, pp. 115-124.

[62] J. Wiebe, T. Wilson, C. Cardie, Annotating Expressions of Opinions and Emotions in Language, Language Resources and Evaluation, 39(2), 2005, pp. 165-210.

[63] T. Nakagawa, K. Inui, S. Kurohashi, Dependency Tree-based Sentiment Classification using CRFs with Hidden Variables, Proceedings of ACL:HLT, 2010, pp. 786-794.

[64] S. Wang, C. Manning, Baselines and bigrams: Simple, good sentiment and topic classication, Proceedings of the ACL, 2012, pp. 90-94.

[65] S. Wang, C. Manning, Fast dropout training, In Proceedings of the 30th International Conference on Machine Learning (ICML), 28, 2013, pp. 118-126.

[66] T. Chen, R. Xu, Y. He, X. Wang, Improving sentiment analysis via sentence type classification using BiLSTM-CRF and CNN, Expert Syst. Appl., 72(C), 2017, pp. 221-230.